\documentclass[letterpaper, 10 pt, conference]{ieeeconf}  

\IEEEoverridecommandlockouts
\usepackage[vlined, ruled, boxed, linesnumbered]{algorithm2e}

\SetCommentSty{mycommfont}

\SetKwProg{Function}{}{}{}

\usepackage{graphicx,subfig}
\usepackage{graphics}
\usepackage{amssymb,amsmath}
\usepackage{amssymb}
\usepackage[noend]{algpseudocode}
\usepackage[font={small}]{caption}

\usepackage[usenames, dvipsnames]{color}

\usepackage[normalem]{ulem}

\usepackage{tikz}
\usepackage{textcomp}
\usepackage{lipsum}
\usepackage[noadjust]{cite}
\usepackage{gensymb}
\usepackage{hyperref}

\usepackage{booktabs}

\newcommand{\Tbf}{\mathbf{T}}

\newcommand{\Ical}{\mathcal{I}}

\newcommand{\Lcal}{\mathcal{L}}

\title{\LARGE \bf
Predictive 3D Sonar Mapping of Underwater Environments\\ via Object-specific Bayesian Inference 
}

\author{John McConnell and Brendan Englot
\thanks{J. McConnell and B. Englot are with the Department of Mechanical Engineering, Stevens Institute of Technology,
        Hoboken, NJ, 07030, USA
        {\tt\small$\{$jmcconn1,benglot$\}$@stevens.edu}}%
}

\begin{document}

\maketitle
\thispagestyle{empty}
\pagestyle{empty}

\begin{abstract}
Recent work has achieved dense 3D reconstruction with wide-aperture imaging sonar using a stereo pair of orthogonally oriented sonars. This allows each sonar to observe a spatial dimension that the other is missing, without requiring any prior assumptions about scene geometry. However, this is achieved only in a small region with overlapping fields-of-view, leaving large regions of sonar image observations with an unknown elevation angle. 
Our work aims to achieve large-scale 3D reconstruction more efficiently using this sensor arrangement.
We propose dividing the world into semantic classes to exploit the presence of repeating structures in the subsea environment. We use a Bayesian inference framework to build an understanding of each object class's geometry when 3D information is available from the orthogonal sonar fusion system, and when the elevation angle of our returns is unknown, our framework is used to infer unknown 3D structure. We quantitatively validate our method in a simulation and use data collected from a real outdoor littoral environment to demonstrate the efficacy of our framework in the field. Video attachment: 
\url{https://www.youtube.com/watch?v=WouCrY9eK4o&t=75s}
\end{abstract}

\vspace{-1.5mm}

\section{Introduction}
\begin{figure}[t]
\centering
\subfloat[BlueROV2 with Oculus wide-aperture imaging sonars\label{fig:leading_1}]{\includegraphics[width=0.4\linewidth]{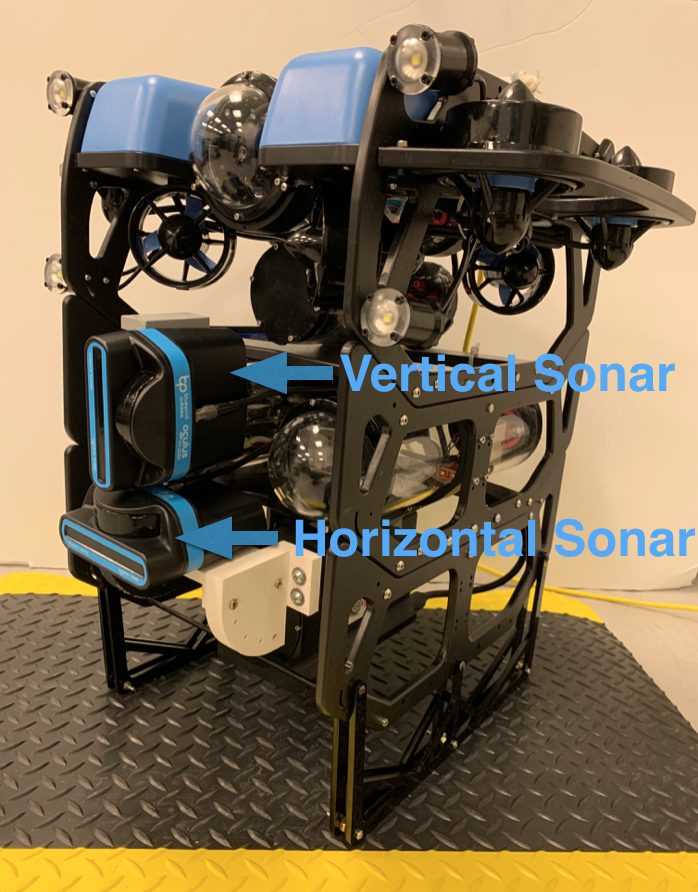}}\ \;
\subfloat[3D Reconstruction w/o and w/ Bayesian inference\label{fig:leading_2}]{\includegraphics[width=0.35\linewidth]{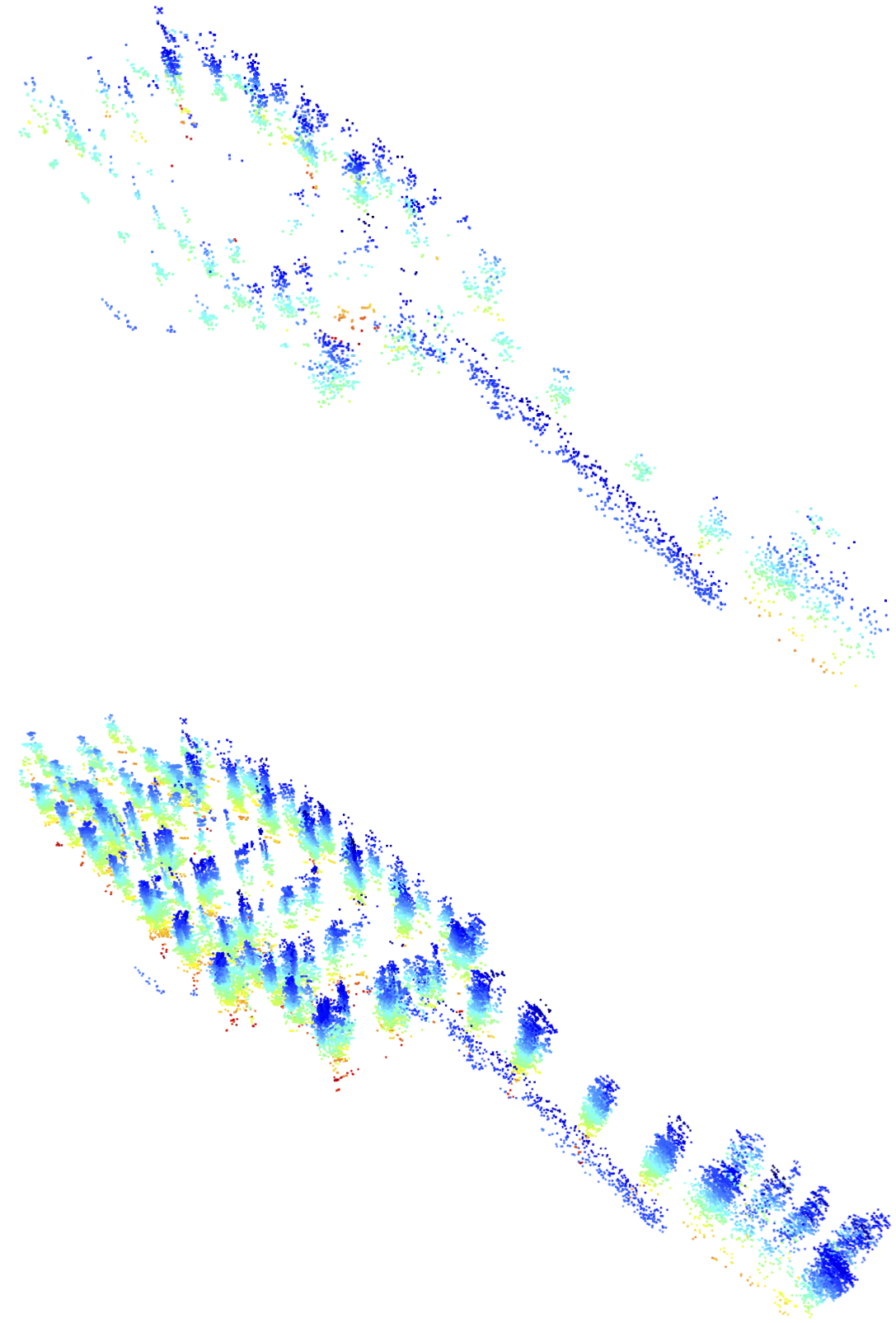}}\\
\subfloat[Satellite image overlay of 3D reconstruction \label{fig:leading_3} ]{\includegraphics[width=0.65\linewidth]{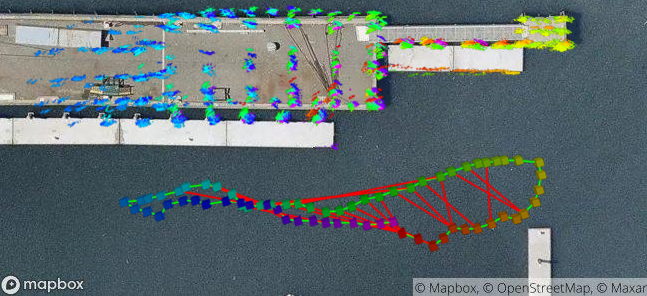}}\
\caption{\textbf{System Overview:} (a) shows our AUV hardware, (b) shows a sample reconstruction from SUNY Maritime's marina in the East River (raw 3D observations at top and with Bayesian inference at bottom) and (c) shows a top-down view of (b) with a satellite image \cite{google} overlay, and the corresponding SLAM pose graph. Squares in (c) show AUV poses with color corresponding to time; green lines are sequential factors; red lines are loop closures. Note: The point cloud in (c) uses the same time mapping as the corresponding poses, while the point cloud in (b) has color mapped to the vertical axis.}
\vspace{-7mm}
\label{fig:leading}
\end{figure}

\vspace{-1mm}

Autonomous underwater vehicles (AUVs) offer important capabilities that support subsea inspection, surveillance, and environmental monitoring. Due to variable water clarity and frequently poor ambient lighting, AUVs often rely on acoustic rather than optical sensors for perception in real-world operational settings; among the most capable and versatile acoustic sensors are multi-beam profiling sonars and wide aperture imaging sonars. Profiling sonars are highly accurate, but only provide a narrow vertical beam. This narrow beam limits the volume of water which can be imaged at each time step, making large-scale 3D mapping with these sensors a prolonged endeavor. Further, using these sensors may be prohibitively expensive, especially considering the high-grade inertial navigation systems (INS) often required to support 3D mapping. Wide aperture, multi-beam imaging sonar, in contrast, maximizes situational awareness by imaging a large volume of water at every time step. However, while these sensors image large 3D volumes, only the range and bearing of their returns are recorded; not the elevation angle.

Recent work \cite{McConnell-2020, Negahdaripour-2018, Negahdaripour-2020} has investigated 
how to supply the missing information using a stereo pair of wide aperture multi-beam imaging sonars to \textit{measure} the unknown elevation angle rather than estimate it. This methodology requires no prior assumptions about the geometry of the 
objects in view, unlike other algorithms for recovering elevation angle from a single sonar image. In the specific method implemented
in our prior work \cite{McConnell-2020}, 
the need for overlapping views across orthogonally oriented sonars means that 3D volumetric sensing can only be performed across a 20\degree-by-20\degree \ field of view (the horizontal field of view of each sensor would normally be 130\degree).  
Using sensors with only 20\degree \ horizontal and vertical fields of view proves tedious when  mapping a large-scale outdoor environment. The question then remains: how can accurate, large-scale 3D volumetric mapping be achieved efficiently by sonar-equipped AUVs in cluttered environments? 

In this paper, we consider the mapping of shallow water, littoral settings often characterized by repeating objects, such as 
pier pilings, throughout the environment. We will exploit this repeating structure in conjunction with the previously proposed orthogonal stereo sonar fusion system to accelerate 3D reconstruction of large scale littoral environments. The contributions of this paper are as follows: 
\begin{itemize}
  \item A framework to identify and exploit the observation of repeating objects in the environment, using past 3D measurements to infer unknown 3D structure when instances of the same class are partially re-observed.
  \item A simulation study demonstrating the accuracy of the proposed Bayesian inference framework in predicting the 3D structure of partially observed objects.
  \item Real-world experiments that show the efficacy of this framework in mapping large-scale littoral environments, as well as its compatibility with sonar-based simultaneous localization and mapping (SLAM) -- an illustrative overview is provided in Figure \ref{fig:leading}. 
\end{itemize}
First, we will discuss related work and precisely define the problem we aim to solve. Next, we will present the developed algorithm in detail. Finally, we will provide experimental results that confirm our algorithm's utility, both in simulation and using real data from the East River in the Bronx, NY.

\vspace{-1.5mm}

\section{Related Work}
\subsection{3D Reconstruction with Wide Aperture Sonar}
Wide aperture multi-beam imaging sonar is a relatively low cost option for underwater perception which provides a large field of view. However, it can only measure two of three dimensions in the spherical coordinate frame: range and bearing, leaving the third, elevation angle, unknown. This limitation
is addressed in  
\cite{Aykin-2013}, where 3D reconstruction is achieved by estimating each sonar contact's unknown elevation angle. This approach is built upon by \cite{Westman-2019}, which accurately maps the surfaces of concrete pier pilings. However, both studies assume that elevation angle monotonically increases or decreases with range, limiting their application to environments where this assumption holds. Further, both methods require the sonar to be positioned at a downward grazing angle; in shallow, littoral environments, this may create difficulty when trying to distinguish between the structures of interest and returns from the seafloor. 

\vspace{-1.5mm}

\subsection{Space Carving}
Another set of 3D reconstruction methods used to address these issues is space carving. Typically, low-intensity background pixels are used to remove sections of a voxel grid, using multiple views of an object \cite{Aykin-2015, Aykin-2016, Guerneve-2018}; with \cite{Westman-2020} being a recent innovation. These methods prove effective in their respective validations but fail to address two fundamental requirements for large scale mapping: the large memory required for such a voxel grid needed to map a large scale outdoor environment, and the accurate online pose estimates required to support this approach. Underwater SLAM is often solved incrementally, with pose estimates fluctuating as the solution evolves. This makes space carving a challenge to implement and a research area unto itself. 

\vspace{-1.5mm}

\subsection{Deep Learning}
Recently, deep learning was applied to this problem in Elevate-net \cite{DeBortoli-2019} by making use of a convolutional neural network (CNN) to estimate the elevation angle at each pixel in a wide aperture sonar's acoustic image. While an exciting line of inquiry, Elevate-net requires pre-training on synthetic data generated with CAD models.

\subsection{Using a Stereo Pair of Sonars for 3D Reconstruction}
Using a stereo pair of imaging sonars to address each sonar's missing elevation angle requires no prior assumptions about object geometry, while providing fully defined 3D points at every time step \cite{McConnell-2020, Negahdaripour-2018, Negahdaripour-2020}. Moreover, because the resulting points are generated independently, one pair of images at a time, 
this approach is easily integrated with a variety of SLAM solutions. 
In the method proposed in our prior work \cite{McConnell-2020}, the use of orthogonally oriented sonars as depicted in Fig. \ref{fig:leading} enables dense mapping, but constrains 3D volumetric sensing to a 20\degree-by-20\degree \ field of view (when the horizontal aperture of each sensor would normally be 130\degree), making large scale 3D mapping a time consuming process.

\subsection{Inference Aided 3D Reconstruction and Mapping}
There is a large body of work that applies probabilistic inference to enhance 3D mapping and reconstruction. Most closely related are the methods that use inference to improve point cloud or voxel mapping, which often use insights from large data sets or object models provided a priori, which is not always available in an underwater robotics setting.  

The use of a variational auto-encoder to infer the 3D distribution of an object given a single view from an RGB camera is explored in \cite{Yu-2018, Yu-2019}. As mentioned above, this work requires a large data set to pre-train the network weights. \cite{Yang-2019} explores the use of synthetic data from a generative adversarial network to avoid this requirement. \cite{Yang-2019} uses a single view voxel grid as the input for 3D reconstruction; the focus is 3D-to-3D densification, rather than the 2D-to-3D inference considered in our work. A similar concept to our work is explored in \cite{Moreno-2013}, where mapping is performed at an object level. Here, object models are produced using high-quality depth camera scans from a controlled setting. These scans are used to improve a 6-DoF SLAM solution. Similar to \cite{Yu-2018, Yu-2019}, the applications of this approach in underwater robotics are limited due to the need for object models a priori. A notable use of inference in an underwater setting is \cite{Guerneve-2017}, where CAD models of objects of interest are provided a priori, and are used to improve the map output. 

A related set of methods use probabilistic inference to enhance occupancy grid maps 
\cite{Wang-2016, Doherty-2017, Meadhra-2018, Gan-2020}. While these methods are excellent at improving occupancy map coverage, they solve mapping under sparse inputs by performing gap-filling and semantic inference on a data structure of the same dimensionality as those sparse inputs. In contrast, we focus on 2D-to-3D inference to enhance the dimensionality of inputs that are not directly observed in 3D, whose inference is conditioned on prior 3D observations of the same class.  

\section{Problem Description}
In this work, we consider 3D reconstruction using a pair of orthogonal imaging sonars with an overlapping field of view. A robot visits a series of poses $x_t$, with transformations  $\Tbf \in \mathbb{R}^{4\times 4}$. Each pose has associated observations $z_t$, with two components: horizontal sonar observations $z^h$ and vertical sonar observations $z^v$. Each set of observations is defined as an intensity image in spherical coordinates with range $R\in \mathbb{R}_+$, bearing $\theta\in \Theta$, and elevation $\phi \in \Phi$, with $\Theta,\Phi \subseteq [-\pi,\pi)$, and an associated intensity value $\gamma \in \mathbb{R}_+$. These measurements can be converted to Cartesian space: 
\begin{align}
    \begin{pmatrix}  X \\  Y \\  Z \end{pmatrix} 
    = R\begin{pmatrix} \cos{\phi} \cos{\theta} \\ 
    \cos{\phi}\sin{\theta} \\ 
    \sin{\phi} \end{pmatrix}.
    \label{eq:tx_to_cartesian}
\end{align}
Each recorded measurement $z^h$ and $z^v$ is characterized by an omitted degree-of-freedom (DoF), 
and in the robot frame, due to the orthogonality of the two sonars, these DoFs differ:  
\begin{align}
    z^{h} &= (R^{h},\theta,\gamma^{h})^\top, &z^{v} &= (R^{v},\phi,\gamma^{v})^\top.
    \label{eq:observations}
\end{align}
As in \cite{McConnell-2020}, we associate measurements across concurrent, orthogonal images to fully define the measurements in 3D, yielding Equation \eqref{eq:fused}:
\begin{align}
    {z}^{Fused}=\left(\frac{R^{h}+R^{v}}{2},\theta^{(h)},\phi^{(v)} \right)^\top.
    \label{eq:fused}
\end{align} 
However, this association can only take place within the small region of overlap between the sonars' fields of view, equivalent in size (in both bearing and elevation) to the sonar's vertical beamwidth. This leaves large portions of each image with a missing DoF and, therefore, undefined in 3D. In this work, we wish to perform a 3D reconstruction by mapping the observations into fixed frame $\Ical$ as in Eq. \eqref{eq:map}. 
\begin{align}
    \mathcal{M} = \{\hat{z}^{(\Ical)}  | \hat{z}^{(\Ical)} = \Tbf \hat{z}^{Fused} \ \forall \  \hat{z}\in  \widehat{z}\}
    \label{eq:map}
\end{align}
The information missing from a portion of all images introduces ambiguity into the application of Eqs. \eqref{eq:tx_to_cartesian} and \eqref{eq:map}. The question to be answered in this paper becomes:  How can we leverage the fully defined measurements from Eq. \eqref{eq:fused} to infer the unknown 3D structure of the rest of the imagery, producing a more comprehensive 3D reconstruction?

\section{Proposed Algorithm} 

We propose to exploit the observation of repeating objects throughout the environment, using the 3D observations occasionally obtained for specific objects, to infer the 3D structure when those same objects are re-observed without a known elevation angle. First, we identify objects in the sonar image and provide a semantic class label. Next, we develop a Bayesian inference model for each object class to later query this model for points in the horizontal sonar image corresponding to that same class. Note that we only apply this inference procedure to the horizontal sonar image. Due to the shallow depths at which we operate our vehicle, the vertical image contains fewer meaningful observations of subsea structures and is not subjected to Bayesian inference.

\subsection{SLAM Solution}
In this work, we utilize a pose SLAM formulation to estimate our robot's pose history through time. We restrict our formulation to 3-DoF estimation in the plane to provide an efficient and robust SLAM pipeline that prioritizes the DoFs of greatest uncertainty (surge, sway, and yaw).

Here we formulate SLAM on a factor graph, which is solved with the aid of GTSAM \cite{GTSAM} and iSAM2 \cite{Kaess-2011}. At each keyframe, features are extracted from our robot's horizontal sonar using the method described below in Sec. IV.B, and the unknown angle $\phi$ is assigned as zero. The constraints relating adjacent keyframes are estimated using iterative closest point (ICP) \cite{besl}, with global initialization via consensus set maximization \cite{consensus} - we denote this step sequential scan matching (SSM). 
Loop closures are added by applying the same scan matching process to non-consecutive keyframes, applying ICP to frames within a designated radius of the current keyframe - we denote this step non-sequential scan matching (NSSM). To reject loop closure outliers, we use pairwise consistent measurement set maximization \cite{Mangelson-2018}. Our factor graph is given by
\begin{flalign*}
\mathbf f(\boldsymbol \Theta) = \mathbf  f^{\text{0}}(\boldsymbol \Theta_0) & \prod_i \mathbf f^{\text{O}}_{i}(\boldsymbol \Theta_i) \prod_j \mathbf f^{\text{SSM}}_j(\boldsymbol \Theta_j) \prod_q \mathbf f^{\text{NSSM}}_q(\boldsymbol \Theta_q).
\end{flalign*}
This planar SLAM solution is used to provide estimates of surge, sway and yaw for registering our algorithm outputs to the global frame; the remaining degrees of freedom come from our vehicle's pressure sensor and inertial measurement unit (IMU). Note that we will only analyze the sonar images corresponding to SLAM keyframes to develop our 3D map. We do not analyze images between keyframes, due to the high drift rate of our low-cost dead reckoning system. 

\begin{figure*}[t]
\centering
{\includegraphics[height=2.5cm]{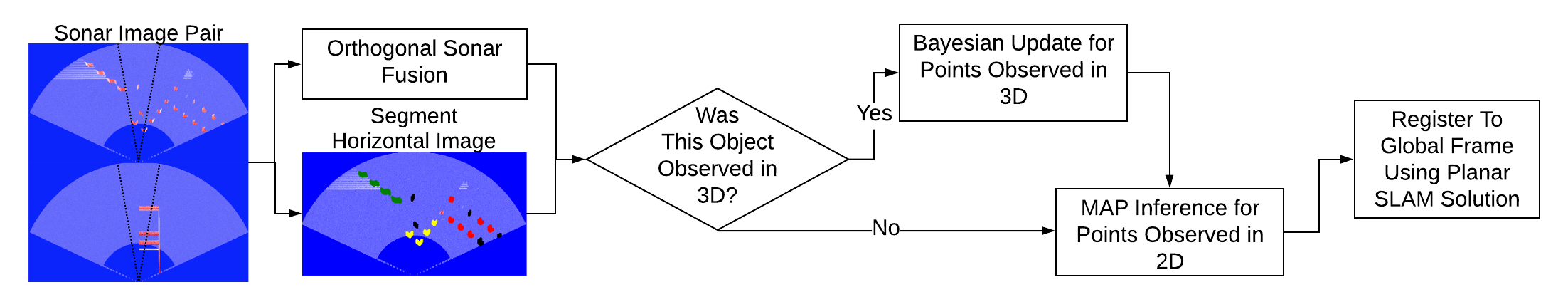}}
\caption{\textbf{System block diagram.} A pair of orthogonal sonar images is provided as input (black lines bound the region of overlap between the two sonar fields-of-view). The images are processed according to Section IV.B. The horizontal image is segmented as in Section IV.C (colors denote different object classes - seawall in green, rectangular pilings in yellow, cylindrical pilings in red). The resulting 3D points enrich each object's model (Eq. \eqref{eq:bayes rule}), while MAP inference is applied to 2D points (Eqs. \eqref{eq:query}, \eqref{eq:query2}).
We then use the planar SLAM solution to register the resulting point cloud. The synthetic sonar images shown here are sampled from the virtual environment depicted in Fig. \ref{fig:gazebo_qual}. }
\vspace{-5mm}
\label{fig:short flow}
\end{figure*}

\subsection{Sonar Fusion System}
At each time step, the robot receives a set of observations constituting a pair of sonar images, as described in Eq. \eqref{eq:observations}. To fully define these in 3D as denoted in Eq. \eqref{eq:fused}, we need to extract and associate features across the images. 

Here we use a similar approach to that of our prior work \cite{McConnell-2020}, dividing the image-feature matching problem into multiple, smaller, subproblems. Recall that range $R$ is discrete within a sonar image, and since associated features should be at the same range, we use the range to define these subproblems. All features at the same range in each sonar image are gathered and processed using intensity-based association. The cost function used here is defined as follows:
\begin{align}
    \Lcal(z_i^{h},z_j^{v}) = || \nu^h - \nu^v||,
    \label{eq:cost}
\end{align}
where $\nu^h$ and $\nu^v$ are square patches of the sonar image around the given feature. These patches have a fixed size of 5x5 pixels in this work. Note that $\nu^v$ is rotated 90 degrees to account for the orthogonality of the images. Moreover, before this comparison is made, the images' intensity values are normalized at every time step. The cost function in Eq. \eqref{eq:cost} is used to find the solution that minimizes the sum of costs between features for each subproblem.

To estimate our confidence in these matches, we compare the two best solutions for each feature association \cite{Hu-Mor-2010}:
\begin{align}
    C = \frac{\Lcal(z_i^{h},z_j^{v})_{min 2}- \Lcal(z_i^{h},z_j^{v})_{min}}{\Sigma_{0,0}^{i,j} \Lcal(z_i^{h},z_j^{v})}.
    \label{eq:uncer}
\end{align}
While we compute subproblem cost totals in order to find sets of associated features in Eq. \eqref{eq:cost}, confidence is evaluated on a feature-wise basis, comparing the costs for each individual feature association made in the given solution. This gives us a simple metric with which to cull uncertain associations. 

\subsection{Image Feature Extraction and Object Classification}
To extract features from sonar imagery, we use the method employed in \cite{McConnell-2020}, used in various radar and sonar applications. The technique, smallest-of cell averaging constant false alarm rate (SOCA-CFAR) detection \cite{Richards-2005}, takes local area averages around the pixel in question and produces a noise estimate. If the signal is greater than a designated threshold, the pixel is identified as an image-feature.

Next, objects must be identified; for this, we utilize DBSCAN \cite{Ester-1996}, since the number of clusters in an image is not known a priori. The results from this step are image-features clustered into objects with unknown class labels. Note that even with a feature detector as robust as SOCA-CFAR, noise is still present. Accordingly, only clusters with $n$ or more  image-features are passed on to the next step. 

Lastly, we must provide a class label for each  object. In this work, we use a simple neural network to perform semantic labeling of object instances. Specifically, we use a CNN that accepts 40x40 pixel sonar image patches in grayscale, with two convolutional layers. Inputs are generated by fitting a bounding box around each object identified in a sonar image and resizing the bounding box into 40x40 pixels, while preserving the object's aspect ratio. 
We utilize Monte-Carlo dropout in this CNN to reject outliers and uncertain classifications by making $m$ predictions for each object. In this way, we can assess the network's confidence in its predictions, as shown in \cite{Loquercio-2020}.  Uncertain predictions are simply provided with a label ``unknown class." An example of this pipeline in action is shown in Fig. \ref{fig:short flow}.

To train this CNN, a small hand-annotated data set of representative sonar imagery is used, which is not included in the sequences used for validation in this work. To generate sufficient training samples to properly train the model, data augmentation is used. We augment our data by applying Gaussian noise, random flips and random rotations.

\subsection{Bayesian Inference for Objects Observed in 3D}

Each detected object in the horizontal sonar image is now represented by a cluster of features with a class label. These features have a range, bearing, and unknown elevation angle. At this step, the dual sonar fusion system provides an elevation angle for a subset of these points, which lie inside the small region with overlapping fields of view. These are the points we concern ourselves with in this subsection. 

We assume objects of the same class will have similar geometries, as is typical in the humanmade littoral environments, populated with piers, used to validate this algorithm.
Semantic classes are defined with this goal in mind, so that objects with similar geometries are grouped together.  

We use a Bayesian inference framework to estimate the conditional 
distribution P(${z_Z^h} | {z_R^h}, {z_\theta^h}$) for each object class incrementally and online. Note that in this process, we estimate Cartesian ${z_Z}$ and not elevation angle. 
${z_Z}$ is a more accurate indicator of the absolute, rather than relative, height at which an object is observed, since in this work we consider scenarios in which our robot maps the environment at fixed depth, employing planar SLAM.
An object's distribution is updated for every measured 3D point per Bayes rule: 
\begin{align}
P({z_Z^h} | {z_R^h}, {z_\theta^h}) = \dfrac{P(  {z_R^h}, {z_\theta^h} | {z_Z^h})  P({z_Z^h})}{P(  {z_R^h}, {z_\theta^h})}.
\label{eq:bayes rule}
\end{align}
Elevation angles measured by the dual sonar fusion system are treated as measurements of ${z_Z^h}$ at the given range and bearing, corrupted with zero-mean Gaussian noise, $\cal{N}(\mu,\,\sigma^{2})$, forming the measurement likelihood $P({z_R^h}, {z_\theta^h} | {z_Z^h})$. The prior, ${P(z_Z^h})$, is simply the existing distribution corresponding to the ${z_R^h}$ and ${z_\theta^h}$ of the newly observed 3D point. We note that these distributions are maintained throughout the whole time-history of the robot's mission, so they incorporate observations from the current frame along with observations from all previous frames. If an update has never been performed, an initial uniform distribution is used. 

At times we may view an object class at different distances and orientations; for this reason, we register each object to a \textit{reference coordinate frame} before we apply Bayes rule in Eq. \eqref{eq:bayes rule}. The first time we see an object, we note its minimum range and median bearing as the reference frame's origin. These object points are then maintained as a ``reference point cloud" to register the object class's future instances to this coordinate frame. When an object is detected and its distribution P(${z_Z^h} | {z_R^h}, {z_\theta^h}$) is updated, the object is first registered to the given object class's reference coordinate frame using ICP. The transformed points are evaluated via Eq. \eqref{eq:bayes rule} and are added to the points used to register future object sightings to the reference frame. Our object-specific distributions P(${z_Z^h} | {z_R^h}, {z_\theta^h}$) will next allow us to predict the height of the sonar returns observed only in 2D.
\vspace{-2mm}
\subsection{Predicting 3D Structure via MAP Estimation}

At each time step, after the process detailed in Section IV.D is completed for the objects measured in 3D, we again consider all objects with class labels comprised of a minimum number of features. We now use the posterior distribution of each object's geometry, $P({z_Z^h} | {z_R^h}, {z_\theta^h})$, to predict the height of all 2D points lacking this information. 

Suppose an object belongs to a class with a posterior updated by at least one application of Eq. \eqref{eq:bayes rule}. In that case, we proceed, first with registration to the object class's reference frame as described above, without adding new points to the reference point cloud. Due to the sonar's ambiguity, it may be that there is more than one true ${z_Z^h}$ for a given range and bearing. For this reason, we break maximum a posteriori (MAP) estimation into two steps, as shown in Eqs. \eqref{eq:query}, \eqref{eq:query2}.
\begin{align}
{z_Z^h}  = argmax  P({z_Z^h} | {z_R^h}, {z_\theta^h}), {z_Z^h} \leq 0 \label{eq:query}\\
{z_Z^h}  = argmax  P({z_Z^h} | {z_R^h}, {z_\theta^h}), {z_Z^h} > 0
\label{eq:query2}
\end{align}
If one or both maxima correspond to confidence exceeding a designated threshold, those values are adopted for inclusion in the robot's map. 
Eq. \eqref{eq:tx_to_cartesian} is solved to provide an output in local Cartesian coordinates, $[X,Y,Z]^T$. This process is completed for all objects in view -- the result is a horizontal sonar image with more observations fully defined in 3D, rather than just the few observations inside the region of dual-sonar overlap. The observations are converted to a point cloud and registered to the global map frame per Eq. \eqref{eq:map}.
\vspace{-1mm}
\section{Experiments and Results}
\vspace{-0.5mm}
\subsection{Hardware Overview}
In order to perform real experiments and derive a simulation environment for this work, we use our customized BlueROV2 heavy-configuration robot, shown in Fig. \ref{fig:leading_1}. This vehicle is equipped with an on-board Pixhawk, Raspberry Pi and NVIDIA Jetson Nano for control and computation. We use a Rowe  SeaPilot  Doppler  velocity  log (DVL), VectorNav VN100 inertial measurement unit (IMU) with integrated Kalman filter, and a Bar30 pressure sensor. For perceptual sensors we use a pair of wide aperture multi-beam imaging sonars, a Blueprint subsea Oculus M750d and M1200d. We use the M750d as our horizontal sonar and the M1200d as the vertical sonar. Note that the entirety of this work takes place with these sonars and their simulated versions operating at a range of 30 meters, with 5cm range resolution. 

In order to manage the BlueROV's sensors, SLAM system, and companion 3D mapping system, we use the Robot Operating System \cite{Quigley-2009}, both for operating the vehicle and for playback of its data. The proposed 3D mapping algorithm is applied to real-time playback of our data using a computer equipped with an NVIDIA Titan RTX GPU and Intel i9 CPU. Due to the keyframe spacing employed within our SLAM framework, our mapping process (which operates only on keyframes) sees a sonar frame-rate that does not exceed 0.4Hz. Note that all experiments take place at a fixed depth. 

\subsection{Simulation Study}
In this section, we utilize Gazebo \cite{gazebo} with the UUV simulator \cite{uuv-sim} to quantitatively validate our method. The UUV simulator provides an implementation of \cite{sonar-sim} to simulate wide aperture multi-beam sonar. 
This simulation environment allows us to perform a quantitative study impossible with field data, where perfect ground truth information is available. We design a humanmade, littoral environment that resembles many marinas and waterfronts. This environment includes two differently shaped pilings (cylindrical and rectangular), boat hulls, corrugated seawalls, and trusses. 

\begin{figure}[t]
\centering
\subfloat[4m Keyframes, Benchmark \label{fig:6a}]{\includegraphics[width=0.4\linewidth]{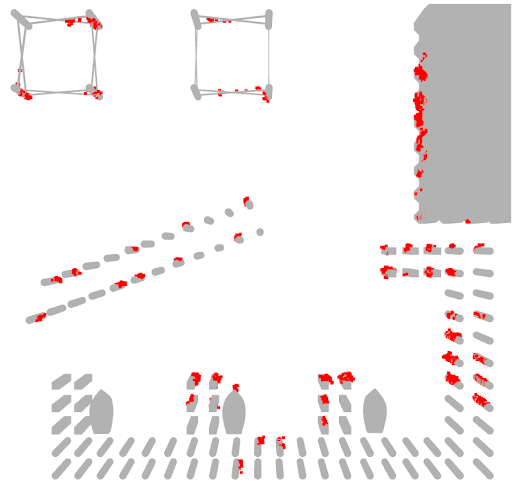}}\
\subfloat[4m Keyframes, Proposed  \label{fig:6a}]{\includegraphics[width=0.4\linewidth]{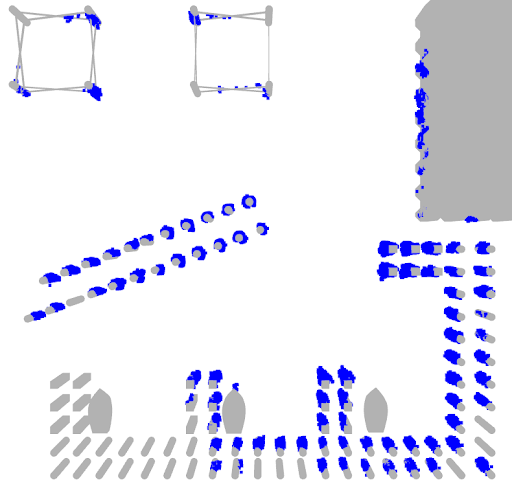}}\vspace{-4mm}\

\subfloat[2m Keyframes, Benchmark \label{fig:6b}]{\includegraphics[width=0.4\linewidth]{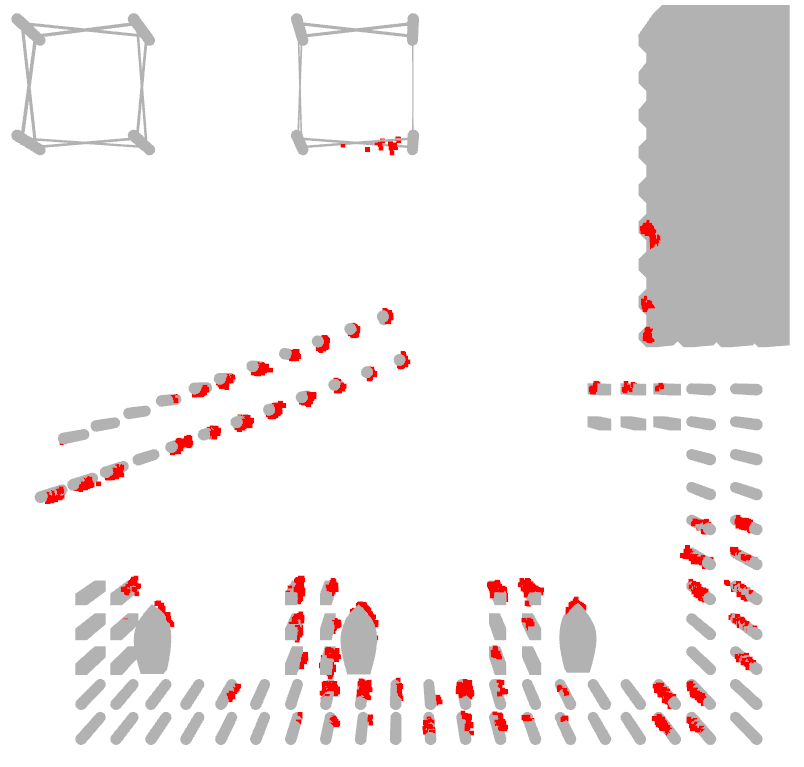}}\
\subfloat[2m Keyframes, Proposed  \label{fig:6c}]{\includegraphics[width=0.4\linewidth]{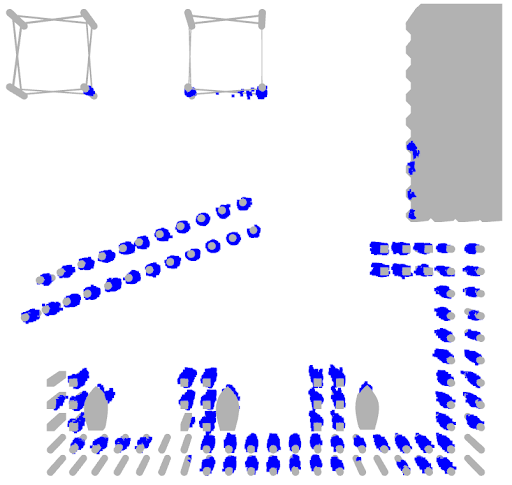}}\vspace{-2mm}\
\vspace{-2mm}
\caption{\textbf{Qualitative Simulation Results.} The top row and bottom row compare results from 4m keyframe spacing and 2m keyframe spacing, respectively. Our proposed method's results are displayed in the right column in blue, and benchmark results (without Bayesian inference) are displayed in the left column in red. 
}
\label{fig:gazebo_qual}
\vspace{-6mm}
\end{figure}

This work proposes an inference method that divides the world into semantic classes by estimating each object class's geometry. These estimates are leveraged to perform 2D-to-3D inference over wide aperture multibeam sonar data. However, not all objects can be treated this way. For example, a wall detected in a sonar image will be difficult to accurately register to a reference coordinate frame, since it may not be fully captured within a single image. Moreover, a pier piling near the edge of a sonar image cannot be distinguished from the edge of a wall that extends beyond the image. Conversely, objects that fit entirely inside the sonar image can be estimated with some confidence, as the registration process described above is readily applicable. It is for this reason that we confine our inference framework in simulation to two classes, cylindrical and rectangular pier pilings. The remaining classes of boat hull, truss and wall, though present in our Gazebo model, are not considered in our predictive mapping framework. A 200-image per class hand-annotated dataset is used to train the CNN.

\begin{figure}[t]
\centering
\includegraphics[width=0.7\columnwidth]{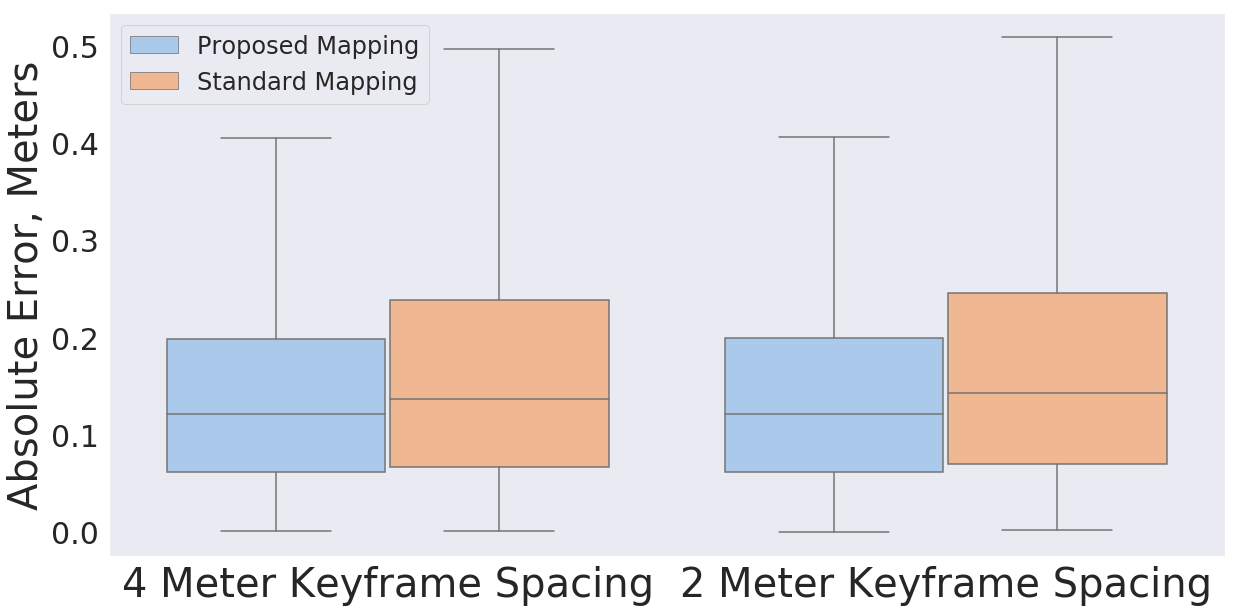}
\caption{\textbf{Simulation Error.} On the left we show the distribution of absolute errors for a keyframe spacing of 4m. On the right we show the distribution of absolute errors for a keyframe spacing of 2m. Errors of our proposed method are displayed in blue; the gold shows errors of simply registering dual sonar fusion outputs. Outliers not shown comprise between 1.7-2.0 percent of each dataset.}

\vspace{-5mm}
\label{fig:box_plots}
\end{figure}

Recall that we use planar pose SLAM to provide transformations to the global map frame. To ensure accurate mapping, only imagery from SLAM keyframes is introduced as input to our framework. We compare two configurations of varying density in this study, a sparse keyframe spacing of 4 meters and a denser keyframe spacing of 2 meters. Note that in this simulation study, while we utilize the same down-sampling as a real SLAM solution, we provide the SLAM pipeline with the robot's true pose in order to isolate errors to only the 3D reconstruction system. As a benchmark for comparison we use a simple method, registering the dual sonar fusion system's 3D output over each keyframe, without classification or Bayesian inference.  While naive, this method represents the state of the art in 3D reconstruction with wide aperture multibeam sonar \cite{McConnell-2020}, while minimizing assumptions with regard to the appearance of the environment. 

For every trial, our robot navigates to each of 20 randomly sampled goals, sequencing them using nearest neighbor; starting at a random location. 
Collision avoidance is achieved using a 2D, in-plane roadmap for navigation purposes only, which is in no way used in our 3D mapping algorithm.  We use A* in conjunction with the roadmap to generate trajectories from one goal to the next. Ten trials are run for each SLAM configuration. 
Fig. \ref{fig:box_plots} shows that our predictive 3D mapping method has comparable error values to the benchmark 3D mapping method. Moreover, per Table \ref{table:sim_voxel_count}, the proposed method provides an order of magnitude greater coverage for each SLAM configuration. Most critically though, our method with half as many keyframes (spaced 4m apart) has an order of magnitude better coverage than the benchmark with twice as many keyframes (2m). Qualitative results are shown in Fig. \ref{fig:gazebo_qual}, which clearly illustrates the improvement in point cloud density over repeating objects, in this case the two different types of pier pilings, without impacting the reconstruction of the trusses, boats, or seawall. 
\begin{table}[h!]
\vspace{-5mm}
\centering
\begin{tabular}{ccc}
\toprule
Algorithm & Mean Voxel Count & Std. Deviation \\
\midrule
Standard Mapping, 4 Meter & 2938.6 & 984.18 \\
\bf Semantic Mapping, 4 Meter & \bf60091.4  & \bf20038.116 \\
\midrule
Standard Mapping, 2 Meter & 3549.4   & 1144.72\\
\bf Semantic Mapping, 2 Meter & \bf68422.2  & \bf25518.37 \\

\toprule
\end{tabular}
\caption{\textbf{Simulation coverage results}, where point clouds are voxelized using a 10cm grid cell resolution.} 

\vspace{-4mm}
\label{table:sim_voxel_count}
\end{table}

We use two metrics to quantify performance: absolute error and voxel count. Absolute error is calculated by finding the shortest distance between a given point in the final point cloud and the CAD model of the environment (Fig. \ref{fig:box_plots}). Voxel count (Table \ref{table:sim_voxel_count}) is calculated by taking the final point cloud and voxelizing it, iterating over all the points and placing them in their respective discrete bins. If a voxel contains one or more points, it is counted; otherwise, it is not. This study uses voxel count to quantify coverage, while not using redundant information contained in the point cloud. 
\subsection{Field Results}

We next apply our method to real sonar data, using two datasets collected in SUNY Maritime College's marina on the East River in Bronx, NY. This environment represents the canonical humanmade littoral environment populated by piers, seawalls, and steel floating docks. Further, the East River poses serious environmental challenges such as low visibility, drastic tidal changes, and currents up to 2 knots. 

To generate data, our BlueROV is manually piloted along the perimeter of structures in the marina while the vehicle heading is held near constant, with the vehicle strafing to either port or starboard. 
In the 300-image, hand-annotated dataset used to train our system, we use two classes: cylindrical pier piling and wall. Due to the aforementioned uncertainty associated with edge features, we only apply our inference method to the cylindrical pier piling class. 

\begin{table}[t]
\vspace{-0mm}
\centering
\begin{tabular}{ccc}
\toprule
& \multicolumn{2}{c}{Voxel Count} \\
Algorithm & 2m Keyframe & 4m Keyframe \\
\midrule
Standard Mapping, Pier & 4147 & 2040\\
\bf Semantic Mapping, Pier & \bf 38143  & \bf 20766 \\
\midrule
Standard Mapping, Waterfront & 5891 & 2819\\
\bf Semantic Mapping, Waterfront & \bf 32131  & \bf 13135 \\
\toprule
\end{tabular}
\caption{\textbf{Field coverage results}, where point clouds are voxelized using a 10cm grid cell resolution. ``Pier" (shown in Fig. \ref{fig:leading}) and ``Waterfront" refer to two large structures in the SUNY Maritime marina, detailed in our video attachment.}
\vspace{-6mm}
\label{table:real_voxel_count}
\end{table}

Since ground truth for mapping is not available, we analyze voxel count to measure coverage and assess the resulting point clouds. 
We produce separate maps for two structures in different areas of the marina, which we term ``Pier'' and ``Waterfront''. 
Once again, our method provides a dramatic increase in coverage, mapping many areas the benchmark leaves blank. Coverage results are shown in Table \ref{table:real_voxel_count}, and detailed maps are provided in our \textcolor{blue}{\href{https://www.youtube.com/watch?v=WouCrY9eK4o&t=75s}{video attachment}}. An overview of the Pier dataset is provided in Fig. \ref{fig:leading}. Roughly 3000 object classifications occur in this dataset.
While we only apply our inference method to one class, because this class is constantly repeated throughout the environment, map coverage is greatly improved. Most critically, these datasets show the utility of our method on real-world sonar data gathered in a complex littoral environment. 

\vspace{-1mm}
\section{Conclusions}
In this paper we have proposed using semantic classes to aid 2D-to-3D Bayesian inference over wide aperture multibeam sonar data. We have shown through experimental validation that exploiting the repeated observation of common structures in a littoral environment can permit highly accurate predictive mapping, without the need for CAD models a priori. In the absence of these structures, this method can still be employed, but with reduced coverage. 

\section*{Acknowledgments}
This research was supported by a grant from Schlumberger Technology Corporation - we thank Arnaud Croux for guidance and support. We also thank Robert Crafa for facilitating access to the SUNY Maritime College marina.

{}

\end{document}